\documentclass{article}

     \usepackage[preprint]{neurips_2020}

\usepackage[utf8]{inputenc} 
\usepackage[T1]{fontenc}    
\usepackage{hyperref}       
\usepackage{url}            
\usepackage{booktabs}       
\usepackage{amsfonts}       
\usepackage{nicefrac}       
\usepackage{microtype}      
\usepackage{graphicx}
\usepackage{subfigure}
\usepackage{booktabs}

\usepackage{multicol}

\usepackage{graphicx}
\usepackage{comment}
\usepackage{amsmath,amssymb} 
\usepackage{color}
\usepackage{mathrsfs}
\usepackage{url}
\usepackage{graphicx}
\usepackage{subfigure}
\usepackage{multirow}
\usepackage{bm}
\usepackage{amssymb}
\usepackage{animate}
\usepackage{multicol}
\usepackage{algorithmic}
\usepackage{algorithm}
\usepackage{amsmath}
\usepackage{caption}

\title{Task-agnostic Temporally Consistent \\ Facial Video Editing}

\author{%
  Meng Cao\thanks{This work is done when Meng Cao works as an intern at Tencent AI Lab} \\
  Peking University\\
  \texttt{mengcao@pku.edu.cn} \\
   \And
   Haozhi Huang \\
   Tencent AI Lab \\
   \texttt{matthzhuang@tencent.com} \\
   \AND
  Hao Wang \\
  Tencent AI Lab \\
  hawelwang@tencent.com \\
  \And
  Xuan Wang \\
  Tencent AI Lab \\
  \texttt{xwang.cv@gmail.com} \\
    \And
  Li Shen \\
  Tencent AI Lab \\
  \texttt{mathshenli@gmail.com} \\
    \And
  Sheng Wang \\
  Peking University \\
  \texttt{shengwangandy@gmail.com} \\
    \And
  Linchao Bao \\
  Tencent AI Lab \\
  \texttt{linchaobao@tencent.com} \\
    \And
  Zhifeng Li  \\
  Tencent AI Lab \\
  \texttt{michaelzfli@tencent.com} \\
      \And
  Jiebo Luo \\
  University of Rochester \\
  \texttt{jluo@cs.rochester.edu} \\
}

\begin{document}

\maketitle

\begin{abstract}
Recent research has witnessed the advances in facial image editing tasks. For video editing, however, previous methods either simply apply transformations frame by frame or utilize multiple frames in a concatenated or iterative fashion, which leads to noticeable visual flickers. In addition, these methods are confined to dealing with one specific task at a time without any extensibility. In this paper, we propose a task-agnostic temporally consistent facial video editing framework. Based on a 3D reconstruction model, our framework is designed to handle several editing tasks in a more unified and disentangled manner. The core design includes a dynamic training sample selection mechanism and a novel 3D temporal loss constraint that fully exploits both image and video datasets and enforces temporal consistency. Compared with the state-of-the-art facial image editing methods, our framework generates video portraits that are more photo-realistic and temporally smooth.
\end{abstract}

\section{Introduction}

\vspace{-5pt}
Facial video editing is an extensively studied task in the industry due to its widespread applications. It includes face swapping~\cite{kim2018deep,olszewski2017realistic,korshunova2017fast,nirkin2018face,jin2017cyclegan,nirkin2019fsgan}, face reenactment~\cite{garrido2014automatic,thies2016face2face,suwajanakorn2017synthesizing,averbuch2017bringing,wiles2018x2face,siarohin2019animating,wu2018reenactgan,zakharov2019few,zhao2017dual}, $etc$. Face swapping replaces one person's identity with another while keeping the original attributes unchanged, and face reenactment focuses on transferring the facial poses and expressions from the others.

Despite tremendous advance in image facial editing, it is still challenging to perform video-level editing because of temporal diversity. Simply applying transformations frame by frame inevitably brings temporal inconsistency, such as visual flickers. To address this issue, several methods have been developed. For face swapping, FSGAN~\cite{nirkin2019fsgan} generates the output in an iterative way. However, reusing the previous frames recursively incurs
error accumulation, which limits its practicality. For face reenactment, \cite{zakharov2019few,wiles2018x2face} regularize temporal relationships implicitly by concatenating multiple frames and feeding them to the networks, which lacks explicit supervision.

In this paper, we propose a \textbf{T}ask-agnostic temporally consistent \textbf{F}acial \textbf{V}ideo \textbf{G}enerative \textbf{A}dversarial \textbf{N}etwok termed as TFVGAN, which utilizes a dynamic training sample selection mechanism and a novel 3D temporal loss. Specifically, several facial video editing tasks are jointly trained in an adversarial way with a mixture of image and video datasets. While image datasets have more varieties in identities, video datasets can provide temporal constraints. A dynamic training sample pick-up strategy is adopted to facilitate the hybrid training process. Moreover, based on 3D reconstruction, we introduce a novel temporal loss using the dense optical flow map via barycentric coordinate interpolations. Both the training strategy and the temporal loss supervision guarantee the high fidelity and temporally consistent outputs.

Meanwhile, as a benefit of introducing 3D reconstruction, TFVGAN obtains more manipulation freedom by disentangling a facial image into pose, expression, and identity coefficients. Therefore, TFVGAN can conduct a novel fully disentangled manipulation task remixing the pose, expression and identity from different video portraits. In contrast, using 2D facial landmark hints~\cite{zakharov2019few} has limited freedom because the pose and expression information in landmarks cannot be decoupled.

\begin{figure}[t]
	\centering
	\includegraphics[height=4.5cm]{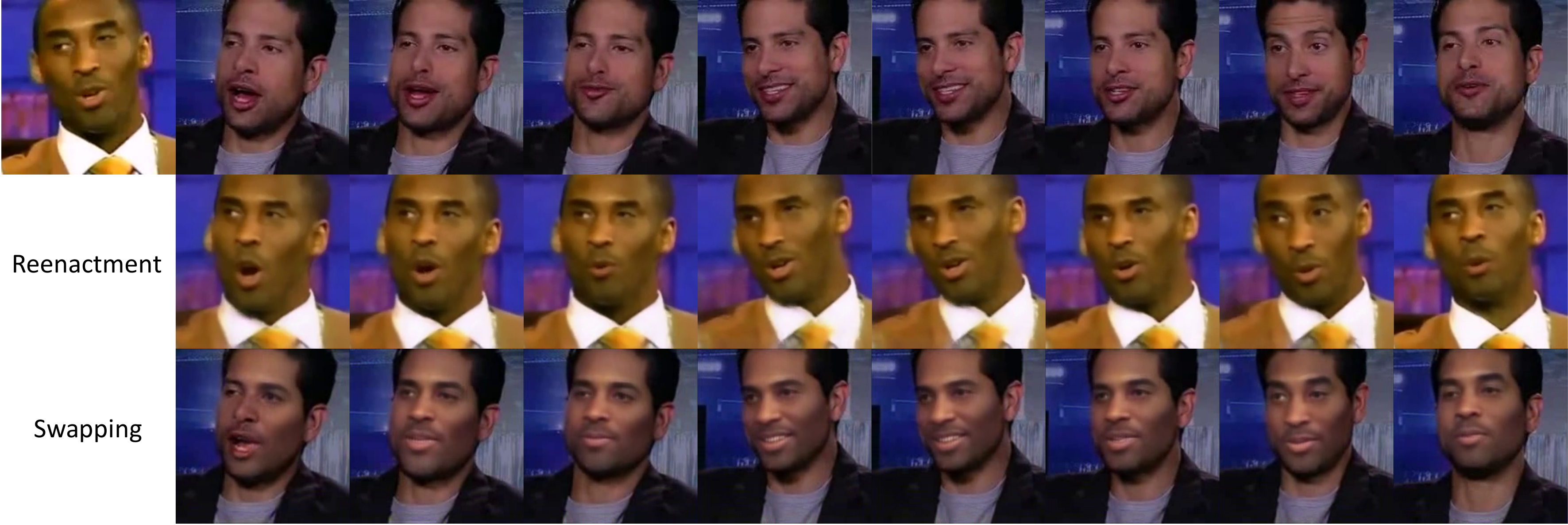}
	\vspace{-5pt}
	\caption{Results of TFVGAN. Face Reenactment and Swapping.}
	\label{fig:teaser}
\end{figure}

TFVGAN has three cascaded procedures. First, we adopt the dynamic selection mechanism to pick up training samples. Next, we incorporate a 3D Morphable Model (3DMM)~\cite{blanz1999morphable}\cite{zhu2016face} reconstruction module to disentangle a video portrait into the pose, expression, and identity coefficients. Afterwards, we recombine these factors according to the specific task to obtain a transformed face sequence through a rasterization renderer. Finally, the rendered transformed face and an auxiliary appearance hint are fused through a deep blending network to generate a final photo-realistic result. 

We conduct extensive experiments to validate our careful designs. Results show that TFVGAN generates impressive outputs with high fidelity and particularly achieves desired temporal consistency on multiple video portrait manipulation tasks.
\vspace{-3pt}
\section{Related Work}
\vspace{-5pt}
\subsection{Facial Image Editing}
\vspace{-5pt}
\noindent\textbf{Face Swapping.} Korshunova et al.~\cite{korshunova2017fast} formulated the face swapping task as a 2D style transfer problem. Jin et al.~\cite{jin2017cyclegan} utilized CycleGAN to transfer facial expressions and head poses more consistently. IPGAN~\cite{bao2018towards} uses two separate encoder networks to achieve the identity and attribute latent vectors and a generator to generate the output based on the concatenation of the two vectors.

\noindent\textbf{Face Reenactment.} X2Face~\cite{wiles2018x2face}  learns an embedded face representation and maps it to the generated frame. ReenactGAN~\cite{wu2018reenactgan} maps the source face onto a boundary latent space and then feeds to a target-specific decoder. MarioNETte~\cite{ha2019marionette} focuses on the identity preserving problem which frequently accrues in face reenactment and proposes a landmark transform technique to produce reenactments of unseen identities.
\vspace{-5pt}
\subsection{Temporal Consistency}
\vspace{-5pt}
Several publications investigate the temporal consistency problem.  FSGAN \cite{nirkin2019fsgan} used an iterative generation way in face reenactment and segmentation stages. \cite{zakharov2019few} and  \cite{ha2019marionette} propose a few-shot neural talking head model based on the concatenation of several frames.

However,  iterative refinement tends to propagate the previous frame artifacts to the next one whereas simple concatenation lacks explicit supervision. In contrast, our proposed TFVGAN explicitly uses a temporal loss to regularize consecutive frames. Besides, a hybrid training dataset with a mixture of image and video data is also adopted to implicitly enforce the  temporal consistency.

\vspace{-5pt}
\section{Proposed Approach}
\vspace{-10pt}
\label{headings}
\begin{figure}[t]
	\centering
	\includegraphics[height=4.5cm]{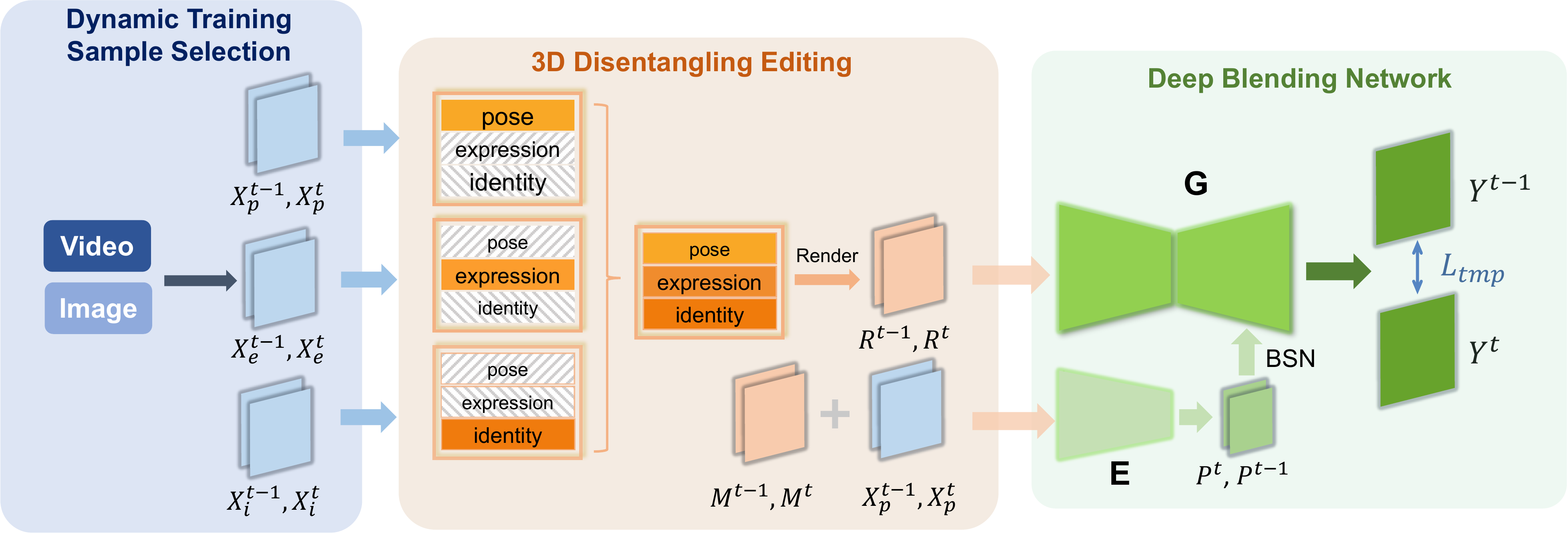}
	\caption{The pipeline of our proposed TFVGAN.}
	\vspace{-5pt}
	\label{fig:pipeline}
\end{figure}
The overall pipeline is shown in Fig.~\ref{fig:pipeline}. $X_i^t$, $X_p^t$ and $X_e^t$ mean the identity image, pose image, and expression image at time $t$, respectively. Firstly, we elaborate on the Dynamic Training Sample Selection mechanism in Section~\ref{subsec:Select}. Then 3D Disentangling Editing is conducted in Section~\ref{subsec:3D}. Finally, we present the deep blending network together with the loss function design in Section~\ref{subsec:GAN}.\subsection{Dynamic Training Sample Selection}\label{subsec:Select}
\vspace{-5pt}
Typically, face swapping algorithms use image datasets such as FFHQ~\cite{karras2019style} for training considering the diversity of identities. However, image dataset struggles on the video face swapping task since it provides little temporal information. Besides, since the source and target images are sampled as two different people, it means that we have no ground truth in this case. For face reenactment, training pairs are taken from video datasets such as VoxCeleb2~\cite{chung2018voxceleb2}, usually sampled as consecutive frames from one individual video to form a self-supervised learning scenario.

In this paper, we demonstrate a joint training of face swapping and face reenactment with a novel dynamic training sample selection mechanism. Specifically, we set the sampling percentage of the image dataset to be $\sigma \in [0,1]$. For video versions, three pairs of consecutive frames ($X_i^t, X_i^{t-1}$), ($X_p^t, X_p^{t-1}$) and ($X_e^t, X_e^{t-1}$) are sampled from one individual video clip. It means that $X_i^t$, $X_p^t$, and $X_e^t$ have the same identity, which forms self-supervision. For image versions,  $X_p^t$, $X_e^t$, $X_i^t$ are randomly sampled from image datasets with different identities. $X_p^t$ and $X_p^{t-1}$ are set to be the same frame, so are the expression images and identity images. In this way, we use video datasets to provide ground truth supervision and image datasets with multiple identities to make the model more robust.

\subsection{3D Disentangling Editing}\label{subsec:3D}
\vspace{-5pt}
Since frames at time $t$ and $t-1$ share the same 3D reconstruction procedures, we ignore the superscript $t$ and $t-1$ here:
Given \(X_i^t\), \(X_p^t\) and \(X_e^t\), we estimate the 3DMM face shapes and the camera projection matrices using 3DDFA~\cite{zhu2017face}. 
Mathematically, the reconstructed 3D face can be expressed as follows:
\begin{align}
S^x = \overline S + A_{\text{id}} \alpha_{\text{id}}^x + A_{\text{exp}} \alpha_{\text{exp}}^x ,
\end{align}
where \(x = i, p, e\) denotes that we are dealing with \(X_i\), \(X_p\) and \(X_e\) respectively. To be specific, \(S^x\) is a 3D face while \(\overline S\) is the mean shape. \(A_{\text{id}}\) and \(A_{\text{exp}}\) are the principle axes derived from BFM~\cite{paysan20093d} and FaceWarehouse~\cite{cao2013facewarehouse}, respectively. At the same time, the 3D reconstruction process also outputs the camera pose $C^x$ (i.e., the position, rotation and scale information). 
Next, we recombine the identity and expression coefficients to acquire the desired 3D face:
\begin{align}
S = \overline S + A_{\text{id}} \alpha_{\text{id}}^i + A_{\text{exp}} \alpha_{\text{exp}}^e,
\end{align}
where \(S\) is the transformed 3D representation for the desired face. 
Meanwhile, we adopt $C^p$ as the desired pose.
Finally, we project the 3D face $S$ with the texture map from $X_i$ onto the 2D image plane to obtain the rendering result:
\begin{align}
R = P(S, T^i, C^p),
\end{align}
where \(P\) is a rasterization renderer with the Weak Perspective Projection according to the camera pose $C^p$, and $T^i$ denotes the texture map from $X_i$. The rendered face image $R$ contains the identity of \(X_i\), the expression of \(X_e\), and the pose of \(X_p\). For more details, please refer to \cite{zhu2017face}. Afterwards, the rendered face image $R$ is used to compute a binary facial region mask $\hat{M}$ with facial areas set to 1 and non-facial areas set to 0. Then we generate the appearance hint $M$. Specifically, for the image case, we set $M = X_p (1 - \hat{M})$ since swapping results require the same appearance with $X_p$; for the video case, to conduct face reenactment, we set $M = X_i (1 - \hat{M})$ for similar reasons.

\subsection{Deep Blending Network}\label{subsec:GAN}
\vspace{-5pt}
In this section, we describe the GAN-based Deep Blending Network which consists of an appearance embedder \(E\), an encoder-decoder structure generator \(G\), and a discriminator \(D\). Generally, for input frames at times $t$ and $t-1$, they share the same generation procedure, and only the time $t$ version is explained in the following.

\noindent\textbf{Appearance Embedder \(\mathbf{E}(X_{p}^t,M^{t})\).}  It takes appearance frame $X_{p}^t$ and its associated appearance hint $M^{t}$, and maps these inputs into a low-resolution feature map denoted as $P^t$. 

\noindent\textbf{Generator \(\mathbf{G}(R^t, P^t, \hat{M}^t)\).} It takes the rendered 3D face $R^t$ and the predicted appearance embedding $P^t$ as input, and synthesizes video frame $Y^t$. We generally build $\mathbf{G}$ using the widely adopted image-to-image translation network \cite{johnson2016perceptual} but replace the instance normalization with a novel bidirectional spatial-aware normalization layer (BSN) which is an enhanced version of AdaIN~\cite{huang2017arbitrary}.

AdaIN~\cite{huang2017arbitrary} aligns the channel-wise mean and variance of two feature maps to conduct style transfer, which means that it equally treats all the elements in the same channel. In our case, however, embedding feature maps $P^t$ is appearance-related and should mainly be applied outside facial areas. Therefore, we take one step further to develop BSN which utilizes the facial region mask $\hat{M}^t$ as a prior prompt. Specifically, for each spatial location in the feature map $X$, we compute the identity transfer term and the feature retention term as follows. For the sake of convenience, we ignore the superscript $t$ here:
\begin{equation}
\begin{split}
\text{BSN}(X, Q, H) & = \underbrace{\alpha \cdot \mathcal{T}(XH, Q\overline{H})  + \beta \cdot \mathcal{T}(X\overline {H}, QH)}_\text{feature transfer term}  \\
& +  \underbrace{ (1 - \alpha) \cdot X H  + (1 - \beta) \cdot X \overline {H}}_\text{feature retention term}  
\end{split}
\end{equation}
and $\mathcal{T}(\cdot, \cdot)$ conducts AdaIN  as:
\begin{align}
\mathcal{T}(A, B) = \sigma(B) \Big(\frac{A - \mu(A)}{\sigma (A)}\Big) + \mu(B), 
\end{align}
where $Q$ is an upsampled version of $P$ and $H$ is a downsampled version of $\hat{M}$. $\overline{H} = 1 - H$ indicates the non-facial area. Two learnable parameter vectors $\alpha$ and $\beta$ are adopted for balance. \(\alpha \in \mathbb{R}^{1\times 1\times c}\) and \(\beta \in \mathbb{R}^{1\times 1\times c} \) share the same channel as input \(X\). Each element \(\alpha_i \in [0,1]\) and \(\beta_i \in [0,1], i \in [0, c-1]\), and the initialization values are set to 0.8 and 0.1, respectively.  

\noindent\textbf{Discriminator \(\mathbf{D}(Y^t, X_i^t)\).} Following pix2pixHD~\cite{wang2018high}, we adopt the same multi-scale discriminator and loss functions, except that we replace the least squared loss term~\cite{mao2017least} by the hinge loss term~\cite{lim2017geometric}.

\subsection{Loss Function}\label{subsec:loss}
\vspace{-8pt}
\noindent\textbf{Appearance Preserving Loss.} The generated facial image $Y^t$ tends to have the same appearance information as $X_p^t$ and the appearance preserving loss is measured in $L_1$ norm as follows:
\begin{align}
\mathcal{L}_{{app}} = || \mathbf{E}(Y^{t})- \mathbf{E}(X_p^t) \|_{1},
\end{align}
where $\mathbf{E}$ is the appearance embedder described in Section~\ref{subsec:GAN}.

\noindent\textbf{Reconstruction Loss.}  When the dynamic training sample selection mechanism chooses a video dataset,  we have ground truths for supervised training since $X_i^t$ and $X_p^t$ are from the same person. Then we have the reconstruction loss as follows:
\begin{equation}
\mathcal{L}_{{rec}} = || Y^{t}- X_i^t \|_{1}.
\end{equation}
Note that for the image case, $\mathcal{L}_{{rec}}$ is set to 0.

\noindent\textbf{Adversarial Loss.} For $\mathcal{L}_{{adv}}$, we use a multi-scale adversarial loss~\cite{wang2018high} on the downsampled output image to enforce photo-realistic results:
\begin{align}
\mathcal{L}_{{adv}} = \frac{1}{K} \sum_{k=1}^{K} \mathbb{E}_{{X_{i,k}^t}}[\log \mathbf{D}(X_{i,k}^t)]+\mathbb{E}_{{Y_k^t}}[\log (1-\mathbf{D}(Y_k^t))],
\end{align}
where $X_{i,k}^t$ and $Y_k^t$ are the downsampled versions of $X_i^t$ and $Y^t$, respectively, and $K$ is the total number of scale versions.
\begin{figure}[t]
	\centering
	\subfigure[]{
		\begin{minipage}[b]{0.5\linewidth}
			\includegraphics[height=0.5\linewidth]{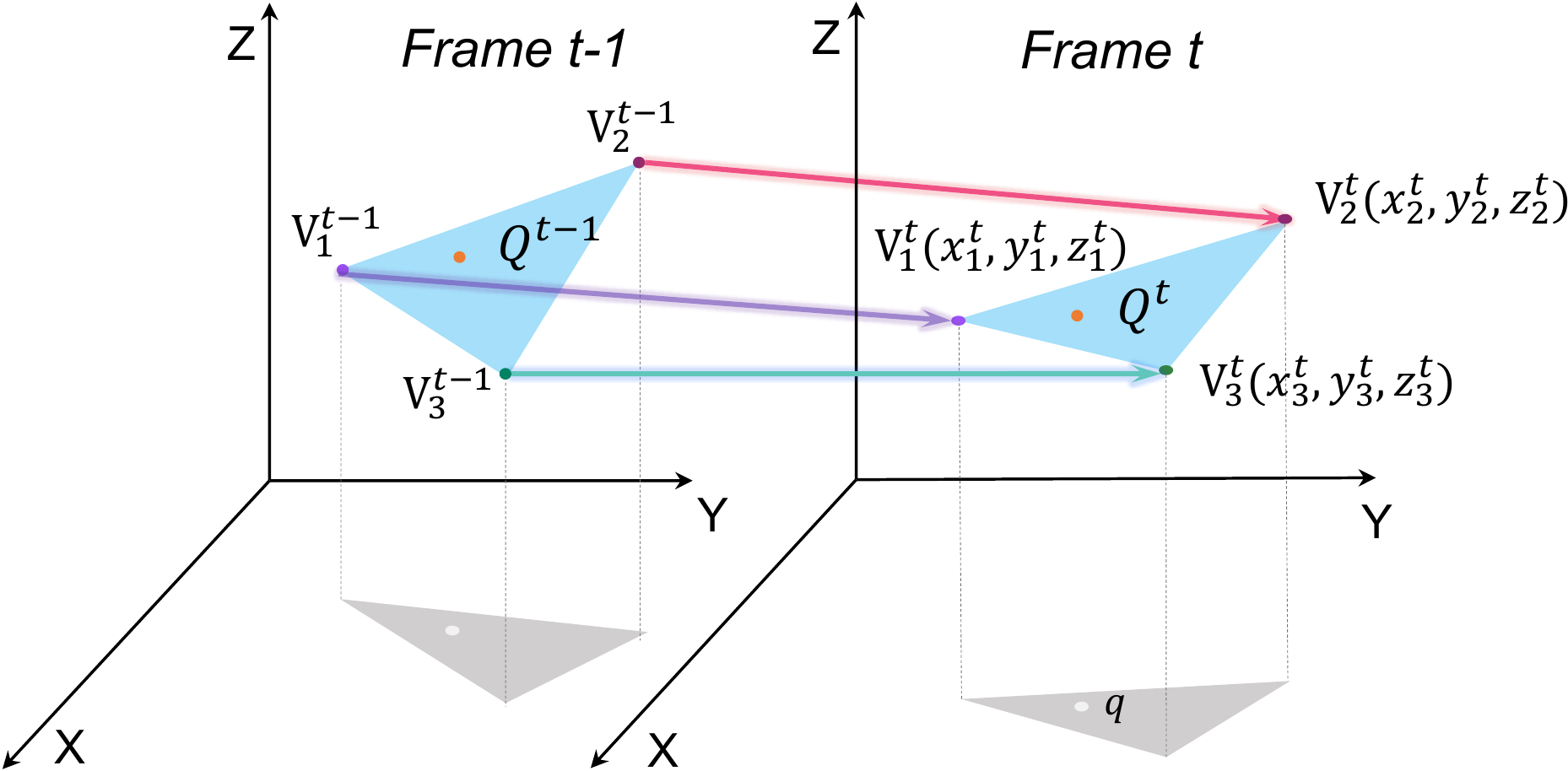}
		\end{minipage}
	}
	\subfigure[]{
		\begin{minipage}[b]{0.4\linewidth}
			\includegraphics[height=0.6\linewidth]{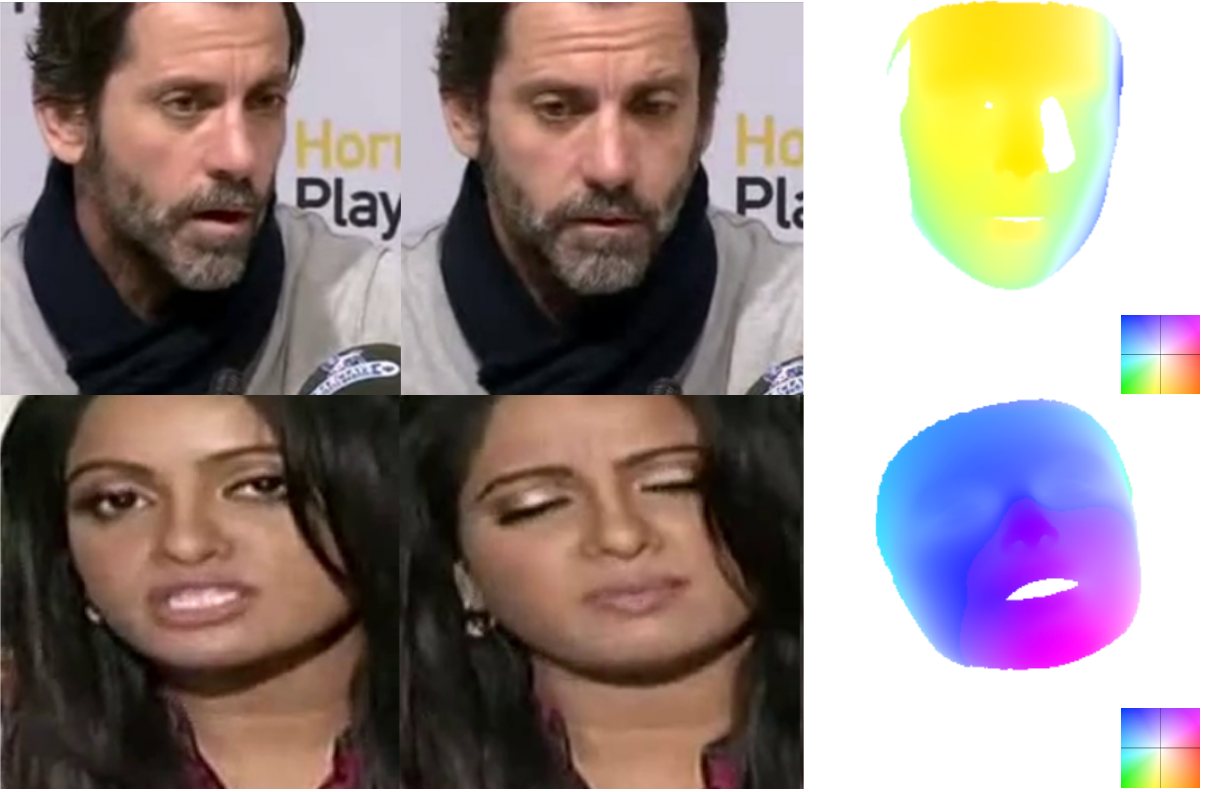}
		\end{minipage}
	}
	\caption{(a) The illustration of barycentric coordinate interpolation of optical flow map. (b) Left: frames at time $t-1$; middle: frames at time $t$; right: generated flow maps $F^{t \rightarrow t-1}$ which warp frame $t$ back to frame $t-1$.}
	\vspace{-5pt}
	\label{fig:flowmap3d}
\end{figure}
\vspace{-5pt}

\noindent\textbf{3D Temporal Loss.} We introduce a flow-based 3D temporal loss to alleviate the inter-frame flicker artifacts. The main difficulty lies in how to generate the dense optical flow map between two consecutive frames. Here, we propose a 3D-based optical flow extraction method, which avoids the use of flow estimation networks~\cite{ilg2017flownet}.

Thanks to 3DMM, the optical flow for reconstructed vertices can be easily obtained by directly subtracting 3D coordinates between adjacent frames. To obtain the dense flow map, we conduct barycentric coordinates interpolation for each pixel in the 2D pixel domain.  As shown in Fig.~\ref{fig:flowmap3d}(a), for a query point $q$ with location $(i,j)$, we find the triangle $T \in \tau$ with its $x$-$y$ plane projection version containing $q$. Let $V_{1}^{t}$, $V_{2}^{t}$, $V_{3}^{t} \in \mathbb{R}^{3}$ be the vertices of $T$. We compute the barycentric coordinates  $\lambda_{1}^{t},\lambda_{2}^{t},\lambda_{3}^{t}$ of $q$, with respect to $T$. Therefore, the flow value at position $(i, j)$ is calculated as follows:
\begin{align}
W_{i, j}^{t \rightarrow t-1}=\sum_{k=1}^{3} \lambda_{k}^{t}\left(V_{k}^{t}-V_{k}^{t-1}\right).
\end{align}
Then we need to obtain visibility maps $S^{t-1}, S^t$. Let $Z^{t-1}$ and $Z^{t}$ be the depth buffer matrices generated by 3DDFA. For frame $t$, we interpolate the depth of $q$ to get the coordinate of $Q$:
\begin{align}
Q^{t} =  (i, j, Q^{t}_z)  = (i, j, \sum_{k=1}^{3} \lambda_{k}^{t} z_{k}^{t}),
\end{align}
where $Q^{t}_z$ denotes the $z$ component of $Q^{t}$. The visibility map $S^t$ value at $(i,j)$ is given as:
\begin{align}
S^t_{i,j}=\left\{\begin{array}{ll}
 1 & \text { if } {Q^t_z} >= Z^{t}_{i,j}\\
0 & \text { otherwise }
\end{array}\right.
\end{align}
where $Z^{t}_{i,j}$ means the depth buffer value of $Z^{t}$ at point $(i,j)$.

For frame $t-1$, we compute the coordinate of $Q^{t-1}$ as follows:
\begin{align}
Q^{t-1} =  ({Q^{t-1}_{x}}, {Q^{t-1}_{y}}, {Q^{t-1}_{z}})  = Q^{t} - W_{i, j}^{t \rightarrow t-1},
\end{align}
where  $Q^{t-1}_{x}$, $Q^{t-1}_{y}$,  $Q^{t-1}_{z}$ are  $x$, $y$, $z$ components of $Q^{t-1}$, and the calculation of visibility map $S^{t-1}$ for frame $t-1$ at position $({Q^{t-1}_{x}, Q^{t-1}_{y}})$  is similar to $S^t$. 
\begin{align}
S^{t-1}_{Q^{t-1}_{x}, Q^{t-1}_{y}}=\left\{\begin{array}{ll}
1 & \text { if } Q^{t-1}_{z} >= Z^{t-1}_{Q^{t-1}_{x}, Q^{t-1}_{y}}\\
0 & \text { otherwise }
\end{array}\right.
\end{align}
Finally, the optical flow map between two consecutive frames is given as follows:
\begin{align}
F_{i, j}^{t \rightarrow t-1} = W_{i, j}^{t \rightarrow t-1} \cdot S^t_{i,j} \cdot S^{t-1}_{Q^{t-1}_{x}, Q^{t-1}_{y}}.
\label{con:flowCal}
\end{align}
Therefore, we define the 3D temporal loss between $Y^t$ and $Y^{t-1}$ in the format of mean squared error:
\begin{align}
\mathcal{L}_{tmp} = || Y^{t-1}-warp\left(Y^{t}, F^{t\rightarrow t-1}\right) \|_{2},
\end{align}
where $warp(Y^{t}, F^{t \rightarrow t-1})$ is the warping function that warps the output at time $t$ back to time $t-1$. You may refer to the algorithm pseudocode in the supplemental material for more details.

The overall loss function is thus reached as :
\begin{align}
 \mathcal{L} =  \arg \min _{\mathbf{G},\mathbf{E}} \max _{\mathbf{D}} \alpha \mathcal{L}_{{adv}}  + \beta \mathcal{L}_{{app}}  + \gamma \mathcal{L}_{{rec}} + \lambda \mathcal{L}_{tmp},
\end{align}
where  $\alpha$, $\beta$,  $\gamma$ and $\lambda$ are the weights to balance different terms, which are set to 10, 1, 10 and 5 in our experiments, respectively.

\section{Experiments}
\vspace{-5pt}
\label{others}

\begin{figure}[t]
	\centering
	\subfigure[]{
		\begin{minipage}[b]{0.14\linewidth}
			\includegraphics[width=1.2\linewidth,height=1.2\linewidth]{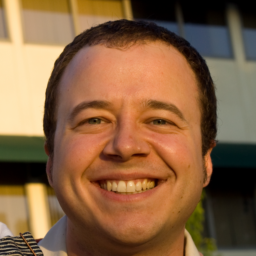}
			\includegraphics[width=1.2\linewidth,height=1.2\linewidth]{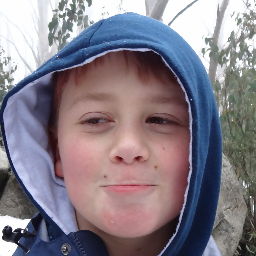}
		\end{minipage}
	}
	\subfigure[]{
		\begin{minipage}[b]{0.14\linewidth}
			\includegraphics[width=1.2\linewidth,height=1.2\linewidth]{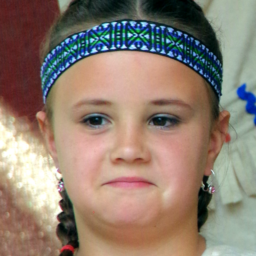}
			\includegraphics[width=1.2\linewidth,height=1.2\linewidth]{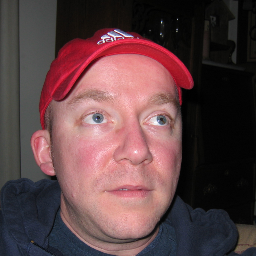}
		\end{minipage}
	}
	\subfigure[]{
		\begin{minipage}[b]{0.14\linewidth}
			\includegraphics[width=1.2\linewidth,height=1.2\linewidth]{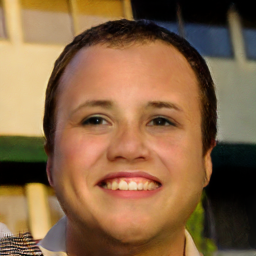}
			\includegraphics[width=1.2\linewidth,height=1.2\linewidth]{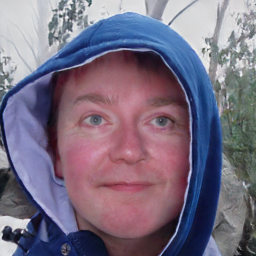}
		\end{minipage}
	} 
	\subfigure[]{
		\begin{minipage}[b]{0.14\linewidth}
			\includegraphics[width=1.2\linewidth,height=1.2\linewidth]{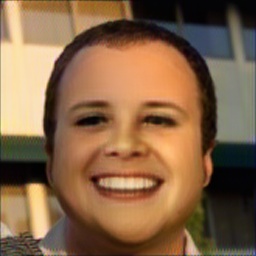}
			\includegraphics[width=1.2\linewidth,height=1.2\linewidth]{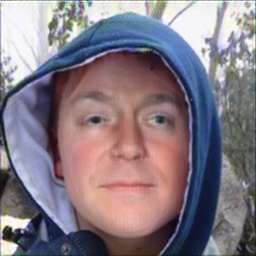}
		\end{minipage}
	}
	\subfigure[]{
		\begin{minipage}[b]{0.14\linewidth}
			\includegraphics[width=1.2\linewidth,height=1.2\linewidth]{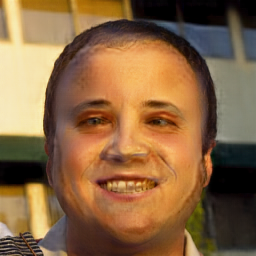}
			\includegraphics[width=1.2\linewidth,height=1.2\linewidth]{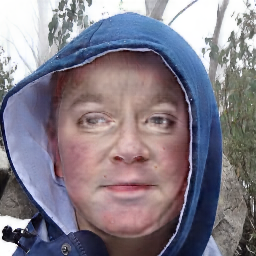}
		\end{minipage}
	}
	\vspace{-5pt}
	\caption{Comparison results of face swapping. (a) Input appearance images $X_p$. (b) Input identity images $X_i$. (c) Results of TFVGAN (Ours). (d) Results of FSGAN. (e) Results of FaceShifter.}
	\vspace{-5pt}
	\label{fig:swapCompare}
\end{figure}

\begin{figure}[t]
	\centering
	\subfigure[]{
		\begin{minipage}[b]{0.14\linewidth}
			\includegraphics[width=1.2\linewidth,height=1.2\linewidth]{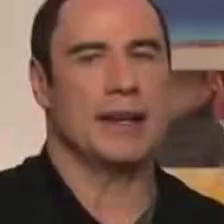}
			\includegraphics[width=1.2\linewidth,height=1.2\linewidth]{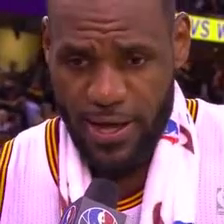}
		\end{minipage}
	}
	\subfigure[]{
		\begin{minipage}[b]{0.14\linewidth}
			\includegraphics[width=1.2\linewidth,height=1.2\linewidth]{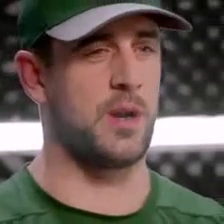}
			\includegraphics[width=1.2\linewidth,height=1.2\linewidth]{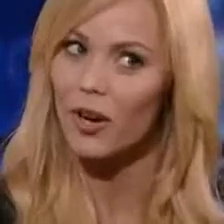}
		\end{minipage}
	}
	\subfigure[]{
		\begin{minipage}[b]{0.14\linewidth}
			\includegraphics[width=1.2\linewidth,height=1.2\linewidth]{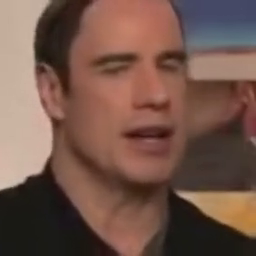}
			\includegraphics[width=1.2\linewidth,height=1.2\linewidth]{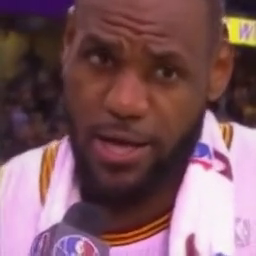}
		\end{minipage}
	} 
	\subfigure[]{
		\begin{minipage}[b]{0.14\linewidth}
			\includegraphics[width=1.2\linewidth,height=1.2\linewidth]{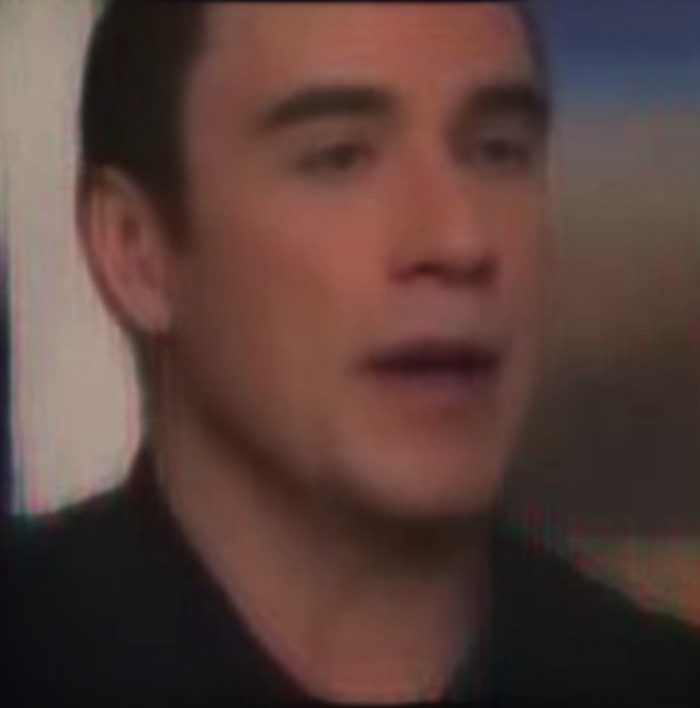}
			\includegraphics[width=1.2\linewidth,height=1.2\linewidth]{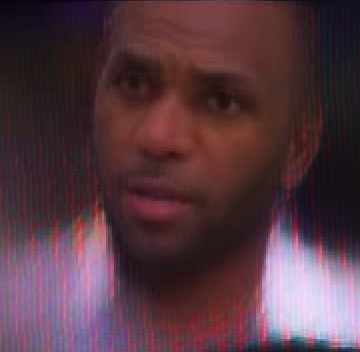}
		\end{minipage}
	}
	\subfigure[]{
		\begin{minipage}[b]{0.14\linewidth}
			\includegraphics[width=1.2\linewidth,height=1.2\linewidth]{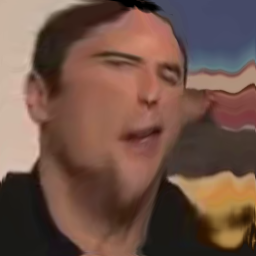}
			\includegraphics[width=1.2\linewidth,height=1.2\linewidth]{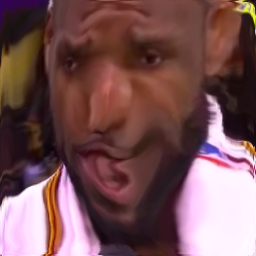}
		\end{minipage}
	}
	\subfigure[]{
		\begin{minipage}[b]{0.14\linewidth}
			\includegraphics[width=1.2\linewidth,height=1.2\linewidth]{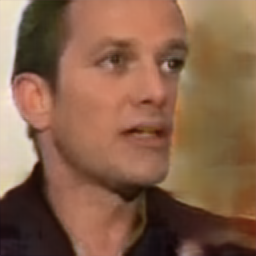}
			\includegraphics[width=1.2\linewidth,height=1.2\linewidth]{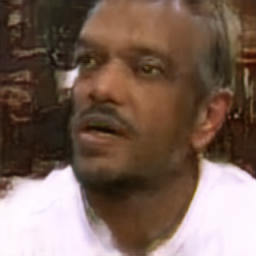}
		\end{minipage}
	}
	\vspace{-5pt}
	\caption{Comparison results of face reenactment. (a) Input identity images. (b) Input pose images. (c) Results of TFVGAN (Ours). (d) Results of FSGAN. (e) Results of X2face. (f) Results of FTH.}
	\vspace{-5pt}
	\label{fig:reenactmentCompare}
\end{figure}

\begin{figure}[t]
	\centering
	\subfigure{
		\begin{minipage}[b]{0.25\linewidth}
			\includegraphics[width=\linewidth,height=\linewidth]{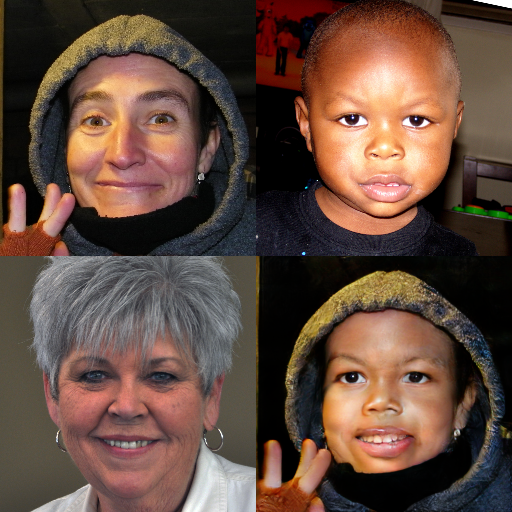}
		\end{minipage}
	}
	\subfigure{
		\begin{minipage}[b]{0.25\linewidth}
			\includegraphics[width=\linewidth,height=\linewidth]{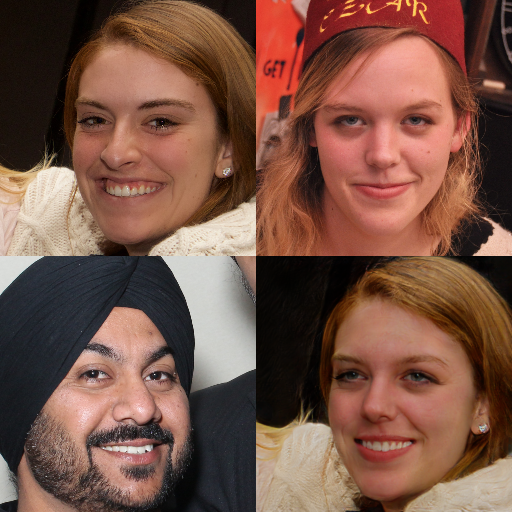}
		\end{minipage}
	}
	\subfigure{
		\begin{minipage}[b]{0.25\linewidth}
			\includegraphics[width=\linewidth,height=\linewidth]{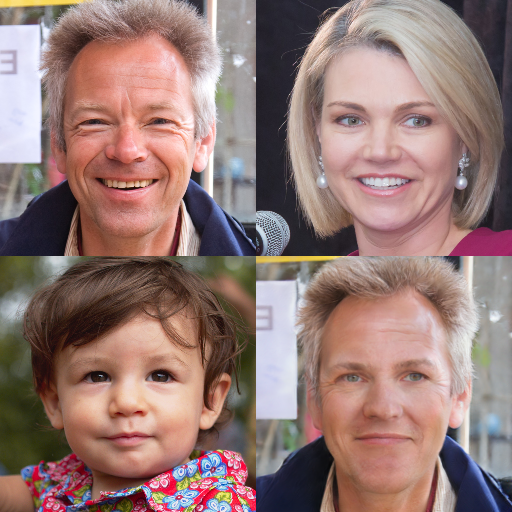}
		\end{minipage}
	}
	\vspace{-5pt}
	\caption{The fully disentangled manipulation task. Top left: appearance images $X_p^t$; top right: identity images $X_i^t$; bottom left: expression images $X_e^t$; bottom right: generated images $Y^t$.}
	\vspace{-7pt}
	\label{fig:faceManipulation}
\end{figure}
In this paper, the FFHQ~\cite{karras2019style} and Voxceleb2~\cite{chung2018voxceleb2} datasets are used as image and video datasets, respectively. FFHQ consists of 70k high-quality images, containing plentiful variations in terms of age, ethnicity and background, while Voxceleb2 covers over a million utterances from over 6,000 speakers. The image dataset ratio $\sigma$ is set to 0.5.

The networks are optimized using Adam optimizer with the learning rate set to \(1 \times 10^{-4}\) for the generator and \(4 \times 10^{-4}\) for the discriminator, respectively. Our framework is implemented using PyTorch and the training is carried out on 8 NVIDIA P40 GPUs with batch size 24.

Quantitative comparisons are conducted with respect to five metrics: Frechet-inception distance (FID)~\cite{heusel2017gans}, structured similarity (SSIM)~\cite{wang2004image}, identity error $E_{\text{id}}$, pose error $E_{\text{pose}}$, and expression error $E_{\text{exp}}$. FID is computed between $Y^t$ and $X_i^t$ while SSIM is only computed for the self-driving face reenactment task. We input $X_i^t$ and $Y^t$ to a face recognition model~\cite{wang2018cosface} to extract identity vectors, and $E_{\text{id}}$ is evaluated in the cosine similarity. To measure pose accuracy, we calculate $E_{\text{pose}}$ using the Euclidean distance between the Euler angles of $X_p^t$ and $Y^t$ extracted by~\cite{ruiz2018fine}. Similarly, $E_{\text{exp}}$ is Euclidean distance between the 2D landmarks of $X_e^t$ and $Y^t$ detected by~\cite{zhang2016joint}.
\vspace{-3pt}
\subsection{Comparison Results}
\vspace{-5pt}
\noindent\textbf{Face swapping.} When we set \(X_e^t = X_p^t\) and crop $M^t$ out of $X_p^t$, it falls into the face swapping problem. Fig.~\ref{fig:swapCompare} presents qualitative results on the FFHQ dataset against FaceShifter~\cite{li2019faceshifter} and FSGAN~\cite{nirkin2019fsgan}. TFVGAN generates high-fidelity results, even under challenging cases with varying ages, poses, gender and hair occlusions. The results of FSGAN are obviously blurred and over-smooth while FaceShiter suffers from striped flaws. Besides, Table~\ref{table:swapReenactCompare} shows that TFVGAN also achieves better quantitative performance, especially in FID.

\noindent\textbf{Face reenactment.}
By setting \(X_e^t = X_p^t\) and cropping $M^t$ out of $X_i^t$ or sampling $X_i^t$ and $X_p^t$ in the same video segments, we deal with the face reenactment task, corresponding to other-driving and self-driving versions, respectively. Table~\ref{table:swapReenactCompare} and Fig~\ref{fig:reenactmentCompare} show the comparison results with state-of-the-art video face reenactment methods X2face~\cite{wiles2018x2face},  FSGAN~\cite{nirkin2019fsgan} and Few-Shot Talking Head (FTH)~\cite{zakharov2019few}. FSGAN and FTH suffer from the problem of background retainment, while X2face shows a distorted face which is a common problem for warping-based methods.

\noindent\textbf{Fully disentangled manipulation.} When $X_i$, $X_p$ and $X_e$ are of different identities, we perform a novel fully disentangled manipulation task. The results in Fig.~\ref{fig:faceManipulation} show that the generated portrait image $Y$ successfully mixes information of identity, appearance and expression from different source images.

\setlength{\tabcolsep}{4pt}
\begin{table}[t]
	\begin{center}
		\caption{Quantitative comparison results on FFHQ for face swapping and on VoxCeleb2 for face reenactment. Note that SSIM for face reenactment is calculated in a self-driving scenario.}
		\label{table:swapReenactCompare}
		\resizebox{0.85\linewidth}{!}{
			\begin{tabular}{ccccccccc}
				\toprule[1pt]
				Mode & Model   & FID \(\downarrow\) & SSIM \(\uparrow\) & $E_{\text{id}}$\(\downarrow\) & $E_{\text{pose}}$\(\downarrow\)  & $E_{\text{exp}}$ \(\downarrow\) \\
				\noalign{\smallskip}
				\toprule[1pt]
				\noalign{\smallskip}
				\multirow{3}*{\scriptsize
					Swapping} & FaceShifter  & 52.1 & - & \textbf{0.15} & (4.31, 3.26, 1.13) & 42.1 \\
				~ & FSGAN & 45.9 & - & 0.74  & (2.27, 3.21, 0.76) & \textbf{24.3} \\
				~ &  TFVGAN & \textbf{34.1} & -   & 0.26 & (\textbf{1.12}, \textbf{1.31}, \textbf{0.65}) & 29.3 \\
				\hline
				\multirow{3}*{\scriptsize
					Reenactment} & X2face  & 52.1 & 0.32  & 0.29 & (2.31, 1.26, 0.9) & 52.4 \\
				~ & FSGAN  & 32.4 & 1.43   & \textbf{0.10}  & (1.43, 0.75, 0.9) & \textbf{35.1}  \\
				~ &  TFVGAN  & \textbf{24.4}  & \textbf{4.93}  & 0.12 & (\textbf{0.21}, \textbf{0.25}, \textbf{0.12}) & 43.1  \\
				\toprule[1pt]
			\end{tabular}
		}
	\end{center}
	\vspace{-3pt}
\end{table}
\vspace{-5pt}
\setlength{\tabcolsep}{1.4pt}

\setlength{\tabcolsep}{4pt}
\begin{table}[t]
	\begin{center}
		\vspace{-5pt}
		\caption{Quantitative results of TFVGAN component ablation on the VoxCeleb2 dataset on face reenactment. (w/o BSN: replacing BSN with AdaIN; w/o $L_{tmp}$: removing temporal loss.) $E_{tmp}$ and SSIM are computed in self-driving face reenactment.}
		\label{table:Ablation}
		\resizebox{0.75\linewidth}{!}{
			\begin{tabular}{cccccccc}
				\toprule[1pt]
				Model   & FID\(\downarrow\) & SSIM\(\uparrow\) & $E_\text{id}$\(\downarrow\) & $E_\text{pose}$\(\downarrow\) & $E_\text{exp}$\(\downarrow\) & $E_\text{tmp}$\(\downarrow\)  \\
				\noalign{\smallskip}
				\toprule[1pt]
				\noalign{\smallskip}
				Full Model  & \textbf{24.4} & \textbf{4.93} & \textbf{0.12} & (\textbf{0.21},\textbf{0.25},\textbf{0.12})  & \textbf{43.1} & \textbf{3.65}\\
				w/o  BSN  & 36.9 & 4.32 & 0.175 & (0.21,0.32,0.12) & 47.2 &-\\
				w/o $L_{tmp}$  & - & -  & - & -  & - & 4.69 \\
				\toprule[1pt]
			\end{tabular}
		}
	\end{center}
	\vspace{-5pt}
\end{table}

\begin{figure} 
	\centering 
	\begin{minipage}[]{.4\textwidth} 
		\centering 
		\includegraphics[height=0.6\linewidth]{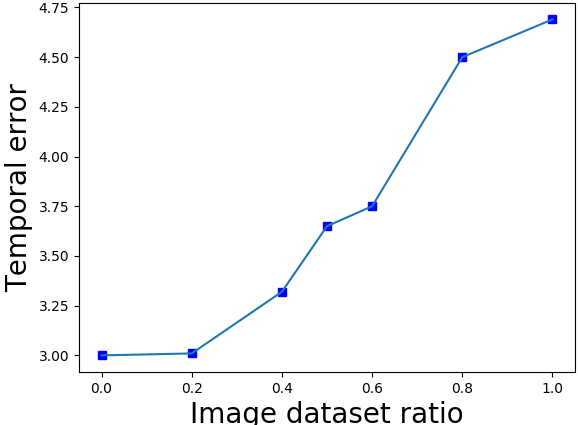}
		\label{fig:DTSS} 
		\caption{Influence of image dataset ratio. Visualization of temporal error $E_\text{tmp}$ \textit{v.s.} image dataset ratio $\sigma$.} 
	\end{minipage}
\begin{minipage}[]{0.5\linewidth}
	\begin{center}
		\label{table:onlyswapReenact}
		\captionof{table}{Influence of joint training. Training jointly  \textit{v.s.} training separately (show as w/o). Note that the required optical flow for computing $E_{\text{tmp}}$ of face swapping is extracted by the TVL1 algorithm in OpenCV since no ground truth is available.}
		\resizebox{0.7\linewidth}{!}{
			\begin{tabular}{cccc}
				\toprule[1pt]
				Mode & Model   & FID$\downarrow$  & $E_\text{tmp}$$\downarrow$\\
				\noalign{\smallskip}
				\toprule[1pt]
				\noalign{\smallskip}
				\multirow{2}*{\scriptsize
					Swapping} & w/o  & 42.5   & 7.98 \\
				~ &  TFVGAN & \textbf{34.1}   & \textbf{3.54}\\
				\hline
				\multirow{2}*{\scriptsize
					Reenactment} & w/o  & 39.9  & 5.32 \\
				~ &  TFVGAN  & \textbf{24.4} &  \textbf{3.65}\\
				\toprule[1pt]
			\end{tabular}
		}
	\end{center}
\end{minipage}
\end{figure}

\setlength{\tabcolsep}{1.4pt}
\vspace{-5pt}
\subsection{Ablation Studies}
\vspace{-5pt}
\noindent\textbf{W and W/O dynamic training sample selection (DTSS).} DTSS takes both image and video datasets as input to achieve high-fidelity and temporally consistent output. To validate its effectiveness,  we keep the network and datasets unchanged and train face swapping and face reenactment alone. Specifically, for face swapping, $X_i$, $X_p$ and $X_e$ are randomly selected from the whole dataset; for face reenactment, image datasets are regarded as single-frame videos. Temporal error $E_{\text{tmp}}$~\cite{huang2017real} is used to evaluate temporal consistency. Table 3 shows the joint training mechanism benefits both tasks. Training alone causes low-quality images (higher FID) and temporal inconsistency (higher $E_\text{tmp}$).

Besides, we shift the image dataset ratio $\sigma$ manually and the results are shown in Fig 7. Intuitively, $E_\text{tmp}$ increases as the image dataset ratio $\sigma$ increases.

\begin{figure}[htb]
	\centering
	\includegraphics[height=3.5cm]{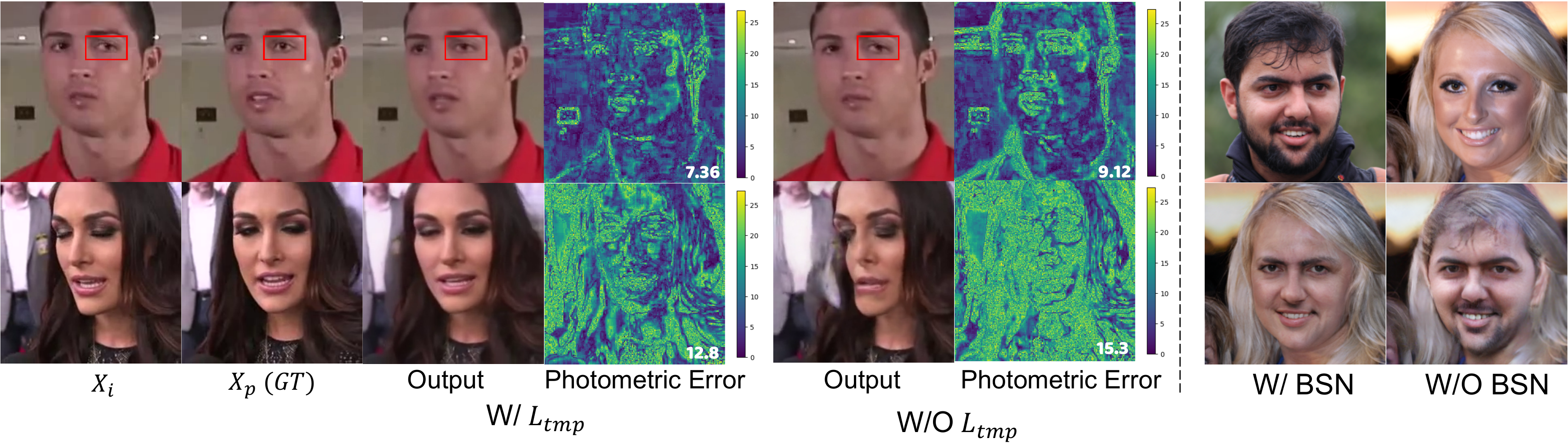}
	\caption{(a) Influence of $L_{tmp}$ on Face Reenactment. $X_p$ is the ground truth because of the self-supervision scenario. The mean photometric error is list bottom-right. Zoom in for a better view and pay attention to the details inside the red boxes. (b) Influence of BSN. Top left: the identity image; top right: the pose image; bottom left: the result with BSN; bottom right: the result without BSN.}
	\label{fig:errormap}
\end{figure}

\noindent\textbf{W and W/O temporal consistency loss.} The ablation study of $L_{tmp}$ is available in Table~\ref{table:Ablation}. Here we evaluate $L_{tmp}$ in the scenario of self-driving reenactment since only in this case we have the ground truth. Several cases are shown in Fig~\ref{fig:errormap} and photometric error~\cite{kim2018deep} is also computed between output and ground truth. Fig~\ref{fig:errormap} (a) shows that $L_{tmp}$ plays an significantly positive role when $X_i$ and $X_p$ have large pose variations. Besides, we also carry out experiments of non-consecutive (long-term) temporal loss and ablation studies of $L_{tmp}$ for face swapping. See supplementary material for details.

\noindent\textbf{BSN vs. AdaIN.} When replacing BSN with AdaIN, the fidelity restoration of TFVGAN degrades substantially. The comparisons between Fig~\ref{fig:errormap} (b) show that BSN leads to a more smooth transition between background and the facial areas and looks more context-harmonious.
\section{Conclusions}
\vspace{-5pt}
In this work, we have proposed a unified framework called TFVGAN for handling multiple video portrait manipulation tasks and generating temporally consistent outputs. To be specific, we carefully design a Dynamic Training Sample Selection mechanism to train face swapping and face reenactment tasks jointly. The whole framework is based on a 3D reconstruction model and a deep blending network. A novel 3D temporal loss is further proposed to enforce the visual consistency in synthesized videos. The extensive experiments have demonstrated the consistent superiority of the proposed framework across multiple tasks over the state-of-the-art methods.

\section*{Broader Impact}
TFVGAN potentially has positive impacts on both academic research and industrial applications. On one hand, TFVGAN can be applied to commercial software to create value. On the other hand, TFVGAN may serve as an assistive technology for a number of related computer vision tasks, such as data augmentation for face recognition, face forgery detection, and so on.

Regarding the possible disclosure of privacy caused by malicious abuse, we will devote ourselves to the face forgery detection task using our proposed TFVGAN. Simply suppressing the publications of face manipulation techniques will not stop their progress, but will hurt the development of face forgery detection.

\small
\bibliographystyle{unsrt} 
\bibliography{refs.bib}

\end{document}